\documentclass{article}
\usepackage[utf8]{inputenc}
\usepackage[english]{babel}

\usepackage[margin=1in]{geometry}
\usepackage{xcolor}
\usepackage{graphicx}
\usepackage{comment}
\usepackage{cite}
\usepackage{authblk}
 
\title{Anomaly Detection using Edge Computing in Video Surveillance System: Review
}
\author[1]{*Devashree R. Patrikar}
\author[2]{Mayur Rajaram Parate}
\affil[1]{*Research Scholar, Indian Institute of Information Technology, Nagpur, India. devashreepatrikar@gmail.com , ORCID id.: 0000-0002-5773-1221}
\affil[2]{Assistant Professor, Indian Institute of Information Technology, Nagpur, India. mparate@iiitn.ac.in, ORCID id.: 0000-0002-2842-6869}
\date{}

\begin{document}

\maketitle

\begin{abstract}

The current concept of Smart Cities influences urban planners and researchers to provide modern, secured and sustainable infrastructure and give a decent quality of life to its residents. To fulfill this need video surveillance cameras have been deployed to enhance the safety and well-being of the citizens. Despite technical developments in modern science, abnormal event detection in surveillance video systems is challenging and requires exhaustive human efforts. In this paper, we surveyed various methodologies developed to detect anomalies in intelligent video surveillance. Firstly, we revisit the surveys on anomaly detection in the last decade. We then present a systematic categorization of methodologies developed for ease of understanding. Considering the notion of anomaly depends on context, we identify different objects-of-interest and publicly available datasets in anomaly detection. Since anomaly detection is considered a time-critical application of computer vision, our emphasis is on anomaly detection using edge devices and approaches explicitly designed for them. Further, we discuss the challenges and opportunities involved in anomaly detection at the edge.
\end{abstract}

\textbf{Keywords- Video Surveillance, Anomaly Detection, Edge Computing, Machine Learning}\\

\textbf{Declarations}
\begin{itemize}

  \item Funding: Not applicable 
  \item Conflicts of interest/Competing interests: Both the authors have checked the manuscript and have agreed to the submission in "International Journal of Multimedia Information Retrieval". There is no conflict of interest between the authors.
  \item Availability of data and material: All the data and material corresponding to the manuscript will be made publicly available. 
  \item Code availability: Not applicable as the manuscript presents a survey of anomaly detection in video surveillance and focus on anomaly detection using edge devices. 
  \item *Corresponding Author: Devashree R. Patrikar, devashreepatrikar@gmail.com, ORCID id.: 0000-0002-5773-1221
\end{itemize}

\newpage

\section{Introduction}
Computer Vision (CV) has evolved as a key technology in the last decade for numerous applications replacing human supervision. CV has the ability to gain a high-level understanding and deriving information by analyzing and processing digital images or videos. These systems are also designed to automate various tasks that the human visual system can do. There are numerous interdisciplinary fields where CV is used; Automatic Inspection, Modelling Objects, Controlling Processes, Navigation, Video Surveillance, etc.
 
'Video Surveillance' is a key application of CV which is used in most public and private places for observation and monitoring. Nowadays intelligent video surveillance systems are used which detect, track and gain a high-level understanding of objects without human supervision. Such intelligent video surveillance systems are used in homes, offices, hospitals, malls, parking areas depending upon the preference of the user. 
 
There are several computer vision-based studies that primarily discuss on aspects such as scene understanding and analysis \cite{wu2019deep} \cite{shirahama2015weakly}, video analysis \cite{long2017edge} \cite{suresha2020study}, anomaly/abnormality detection methods \cite{xu2018anomaly}, human-object detection and tracking \cite{guo2019detecting}, activity recognition \cite{singh2017graph}, recognition of facial expressions \cite{elgarrai2016robust}, urban traffic monitoring \cite{zhou2017fast}, human behavior monitoring \cite{popoola2012video}, detection of unusual events in surveillance scenes \cite{muhammad2019efficient}, etc. Out of these different aspects anomaly detection in video surveillance scenes has been discussed further in our review.
 
Anomaly Detection is a subset of behavior classification. Anomalies are unusual behavior or events that deviate from the normal. Anomaly detection in video scenes is the cutting-edge technology that monitors unusual activities using artificial intelligence. Examples of an anomaly in video surveillance scenes are shown in Figure 1; a person walking in a restricted area, vehicles moving in the wrong direction, a cyclist riding on a footpath; a sudden crowd of people; a person carrying a suspicious bag, a person climbing over the fence, etc.

\begin{figure}
    \centering
   \includegraphics[width=7cm]{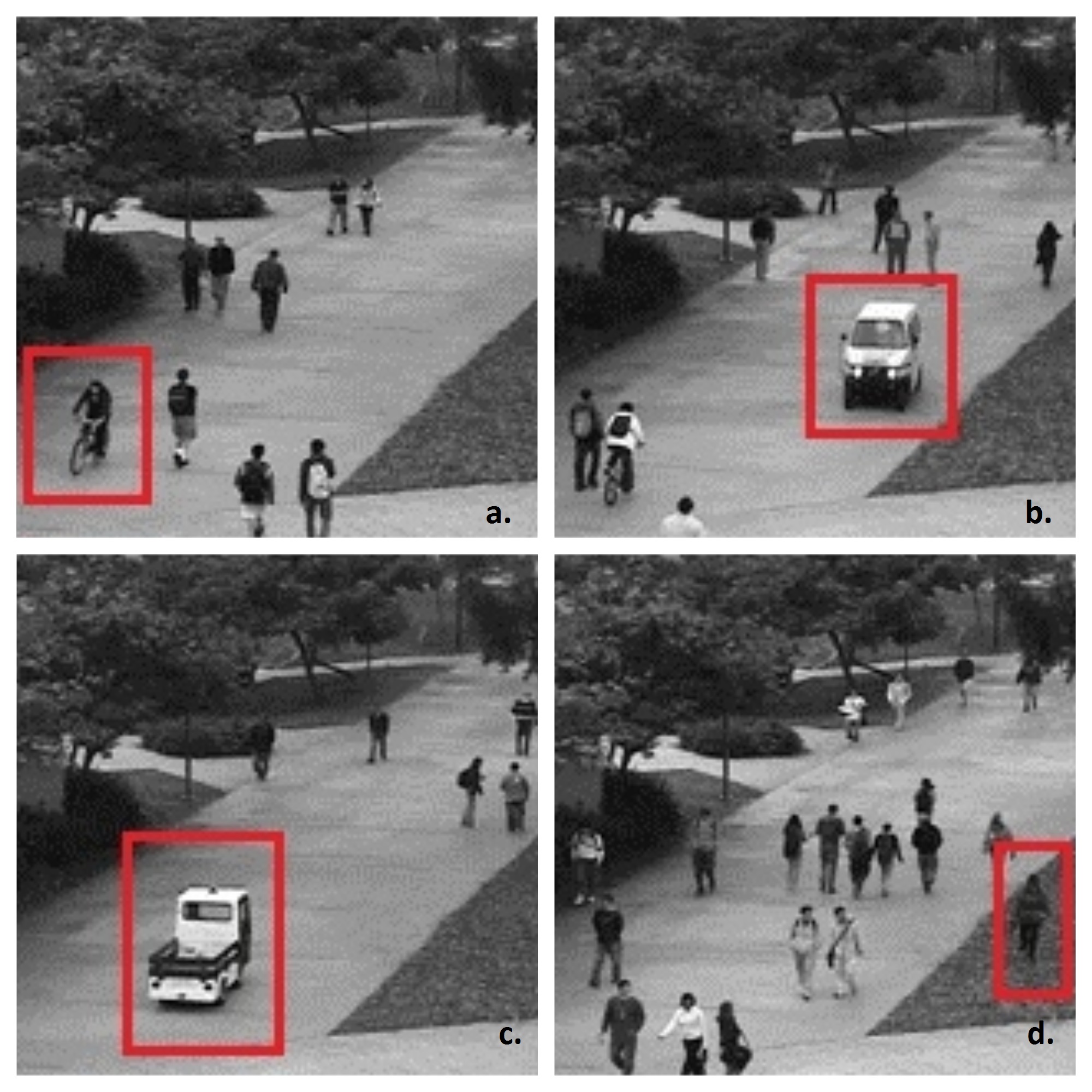}\\
  \caption{Anomaly Detection in video surveillance scenes. Reference: UCSD Dataset. (a) Cyclist Riding on Footpath. (b)(c) Vehicles moving on footpath (d) Pedestrian walking on lawn.} 
\end{figure}

Intelligent video surveillance systems track unusual suspicious behavior and raise alarms without human intervention. Various machine learning tools are used for the detection and tracking of human objects in video scenes and to classify the events as normal and abnormal. The general overview of the anomaly detection is shown in Figure 2. In this process, visual sensors in the surveillance environment collect the data. This raw visual data is then subjected to pre-processing and feature extraction \cite{georgiou2020survey}. The resulting data is provided to a modeling algorithm, in which a learning method is applied to model the behavior of surveillance targets and to determine whether the behavior is abnormal or not. 
For the purpose of anomaly detection, various machine learning tools use cloud computing for data processing and storage \cite{ghosh2020edge}. Cloud computing requires large bandwidth and longer response time \cite{srivastava2016survey} \cite{yang2019edge}. Anomaly detection in video surveillance is a delay sensitive application and requires low latency. So considering these aspects, cloud computing in combination with edge computing provides a better solution for real-time intelligent video surveillance \cite{shi2016edge}. 

The research efforts in anomaly detection for video surveillance are not only scattered in the learning methods but also approaches. Initially, the researchers broadly focused on the use of different handcrafted spatio-temporal features, conventional image processing methods. Recently, more advanced methods like object-level information and machine learning methods for tracking, classification, and clustering have been used to detect anomalies in video scenes. In this survey, we aim to bring together all these methods and approaches to provide a better view of different anomaly detection schemes.

Further, the choice of surveillance target varies according to the application of the system. The reviews done so far have a disparity in the surveillance targets. We have categorized the surveillance targets primarily focusing on four types; automobile, individual, crowd, object, or event.

Moreover, the evolution of cloud and edge devices and their employment in the automated surveillance and anomaly detection is important. Traditionally, massive surveillance data is sent to cloud servers where the data is analyzed and modelled to detect abnormal behaviour or events. With the advancement in the cloud technology, the physical servers is replaced by cloud servers to perform computationally heavy task of computer vision. However, inevitable network latency and operational delays make cloud computing inefficient for the time sensitive applications such as anomaly detection. Thus, this survey discuss the application of Edge Computing (EC) with cloud computing which enhance the response time for anomaly detection. This survey also presents recent techniques in anomaly detection using edge computing in video surveillance.\\

None of the previous surveys address the confluence anomaly detection in video surveillance and edge computing. In this study, we provide a detailed review of latest publications on anomaly detection in video surveillance using edge computing. This review will also address the challenges and opportunities involved in anomaly detection using edge computing.\par  

The research contributions of this review article are as follows:
\begin{enumerate}
    \item Presented review attempts to connect the disparity in the evaluation of the problem formulations and suggested solutions for the anomaly detection. 
    \item The suitability of anomaly detection techniques in the context of application area, surveillance targets, learning methods, and modeling techniques.
    \item  We explore anomaly detection techniques used in vehicle parking, automobile traffic, public places, industrial and home surveillance. The emphasis in these surveillance scenarios is on humans (individual/crowd), objects, automobiles, events and their interactions. 
    \item  The review will also focus on modern-age edge computing technology employed to detect anomalies in video surveillance applications and further discuss the challenges and opportunities involved. 
    
\end{enumerate}

\begin{figure}
    \centering
   \includegraphics[width=10cm]{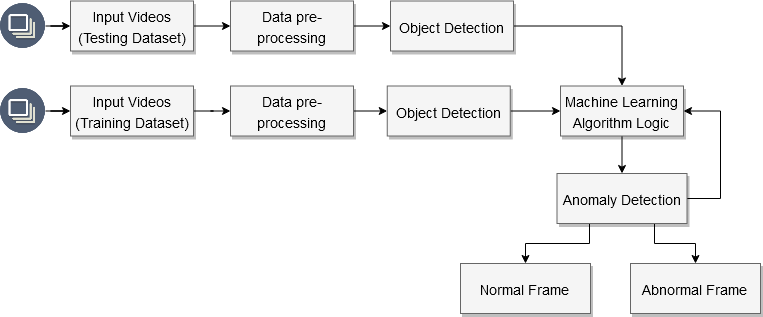}\\
  \caption{General Block Diagram of Anomaly Detection} 
\end{figure}

Further, to the best of our knowledge anomaly detection using edge computing paradigm in video surveillance systems is less explored and not surveyed.\\

We present this survey from the aforementioned perspectives and organize it into seven sections; Section-II presents the prior published surveys, Section-III presents different surveillance targets in corresponding application areas. Section-IV article explores methodologies employed in anomaly detection. Section-V talks about the adoption of edge computing, its challenges, and opportunities in video surveillance and anomaly detection. Section-VI presents critical analysis followed by conclusion in Section-VII

\section{Related Work}
The context of anomaly is different in various scenarios, but can broadly be categorized into anomalies in road traffic and anomalies in human or crowd behavior. Apart from anomalies in human/crowd behavior, on-road vehicle tracking and surveillance are also extensively studied and reviewed over the past decade. The advances in vehicle detection using monocular, stereo vision, and active sensor–vision fusion for an on-road vehicle are surveyed in \cite{sivaraman2013looking}. Automobile detection and tracking are surveyed in \cite{tian2014hierarchical}. The performance dependency of a vehicle surveillance system on traffic conditions is also discussed and a general architecture for the hierarchical and networked vehicle surveillance is presented. The techniques for recognizing vehicles based on attributes such as color, logos, license plates are discussed in \cite{shobha2018review}. The anomaly detection methodologies in road traffic are surveyed in \cite{kumaran2019anomaly}. As the anomaly detection schemes cannot be applied universally across all traffic scenarios, the paper categorizes the methods according to features, object representation, approaches, and models. 

Unlike anomaly detection in-vehicle surveillance, anomalies in human or crowd behavior are much more complex. Approaches to understanding human behavior are surveyed in \cite{popoola2012video} based on human tracking,human-computer interactions, activity tracking, and rehabilitation. In \cite{sodemann2012review}, the learning methods and classification algorithms are discussed considering crowd and individuals as separate surveillance targets to detect the anomaly. However, the occlusions and visual disparity in the crowded scenes reduce the accuracy in detecting the anomalies. A review \cite{li2014crowded} focuses on the aforementioned aspects and learns the motion pattern of the crowd to detect abnormal behaviour. The recognition of complex human behaviour and various anomaly detection techniques are discussed in \cite{li2016anomaly}. Further, the use of moving object trajectory-clustering \cite{yuan2017review}, and trajectory-based surveillance \cite{ahmed2018trajectory} to detect abnormal events are observed in the literature. The recent surveys on anomaly detection and automated video surveillance are listed in Table 1.\\ 

\begin{table}[]
    \centering
    \caption{Recent Surveys on Video Surveillance}
    \label{table_1}
    \begin{tabular}{l|l|l}
        Year & Existing Work & Broad Topics \\
        \hline
        2019 & Ahmed et al. \cite{ahmed2018trajectory} & Trajectory-Based Surveillance \\
         2018 & Shobha et al.\cite{shobha2018review} & Vehicle Detection, Recognition and \\
     & & Tracking \\
     2016 & Yuan et al. \cite{yuan2017review} & Moving object trajectory clustering\\
     2016 & Xiaoli Li et al. \cite{li2016anomaly} & Anomaly Detection Techniques \\

     2015 & Li et al. \cite{li2014crowded} & Crowded Scene Analysis \\
     2014 & Tian et al. \cite{tian2014hierarchical}& Vehicle Surveillance \\
     2013 & Sivaraman et al. \cite{sivaraman2013looking}& Vehicle Detection, Tracking, \\
     2012 & Popoola et al. \cite{popoola2012video} & Abnormal Human Behaviour Recognition \\
     2012 & Angela A. Sodemann \cite{sodemann2012review} & Human Behaviour Detection \\
    \end{tabular}
    
\end{table}

\section{Surveillance Targets}
The entities upon which the surveillance operates are called surveillance targets. Surveillance targets are those entities among which the anomaly detection method aims to detect anomalies. In the context of surveillance areas, the surveillance targets can be categorized as; the individual, crowd, automobile traffic, object or event, the interaction between humans and objects, etc. A Venn diagram showcasing the relationship between video surveillance, anomaly detection, and the surveillance targets (individual, crowd, automobile traffic, object) is illustrated in Figure 3. As shown in Figure 3, there is a large domain of research emphasizing on automated surveillance of targets that are included in the process of anomaly detection and scene understanding.

\subsection{Individual}
Anomaly detection is deployed to ensure safety of individuals in public places like hospitals, offices, public places or at home. It recognizes patterns of human behavior based on sequential actions and detect abnormalities \cite{torres2018detection} \cite{lee2009nonsupervised} . Several approaches have been proposed to detect anomalies in behavior involving breach of security \cite{puvvadi2015cost}, running \cite{zhou2019anomalynet}, lawbreaking actions like robbery \cite{lao2009automatic}, fall of elderly people \cite{liu2021privacy}.

\subsection{Crowd}
This review distinguishes between individuals and crowds as shown in Figure 3. Both of these targets consist of a single entity, the methods used to identify abnormalities are distinct for individuals and crowds \cite{li2013anomaly} \cite{tripathi2019convolutional}. Any change in motion vector or density or kinetic energy indicates an anomalous crowd motion \cite{yuan2014online, kaltsa2015swarm, colque2016histograms, sabokrou2017deep, sabokrou2018deep}. In \cite{li2016anomaly}, behavior such as people suddenly running in different directions or the same direction is considered anomalous. A crowd cannot only be a crowd of individuals but a fleet of taxis as well; \cite{bock2019smart} allows the scene understanding and monitoring on a fleet of taxis. Crowd counting in high density becomes difficult. \cite{khan2020scale} detects only the head of people as it is the only visible part as the body is occluded in public.

\subsection{Automobiles and Traffic}
The automobile and traffic surveillance intends to monitor and understand automobile traffic, traffic density, traffic law violations, safety issues like; accident or parking occupancy. In smart cities, automobiles become important surveillance targets and extensively surveyed for traffic monitoring, lane congestion, and behaviour understanding \cite{zhou2017fast} \cite{kumaran2019anomaly} \cite{bock2019smart} \cite{wang2018offloading} \cite{ke2016real} \cite{ke2018real} \cite{chen2019distributed}. In metro cities, finding a vacant parking spot for vehicles is a tedious job for drivers; \cite{nieto2018automatic} allows drivers to find a vacant parking area. For better accessibility, security, and comfort of the citizens, studies also focus on traffic law violations which include vehicles parked in an incorrect place \cite{ke2020smart}, predicting anomalous driving behavior, abnormal license plate detection \cite{liu2018hybrid}, detection of road accidents \cite{singh2018deep} and detection of collision-prone behavior of vehicles \cite{roy2020detection}.

\subsection{Inanimate objects or events}
The target in this category is divided into events and inanimate objects. Some of the examples of abnormal events are; an outbreak of fire, which is a common calamity in industries \cite{muhammad2019efficient} and needs automatic detection and quick response. Similarly, it is challenging to detect smoke in the foggy environment; \cite{muhammad2019edge} presents smoke detection in such an environment which plays a key role in disaster management. 
Sometimes there are defects in the manufacturing system and it is tedious for humans to examine small details; \cite{li2018deep} proposes a scheme for detecting manufacturing defects in industries.

\subsection{Interaction between humans and objects}
In this category, anomaly detection schemes are associated with the interaction between humans and objects. Both individuals and objects together give the potential benefits of detecting interaction between them such as an individual carrying a suspicious baggage \cite{nawaratne2019spatiotemporal}, individual throwing a chair \cite{angelini2019privacy}. Some studies attempt to account for both pedestrians and vehicles in the same scene such as cyclists driving on a footpath, pedestrians walking on the road \cite{nawaratne2019spatiotemporal} \cite{xu2018real} \cite{wang2019saliencygan} \cite{shojaei2019semi}. In \cite{sabokrou2017deep} abnormal behavior is identified by objects like a skateboarder, a vehicle or a wheelchair on footpath.

\begin{figure}
    \centering
   \includegraphics[width=6cm]{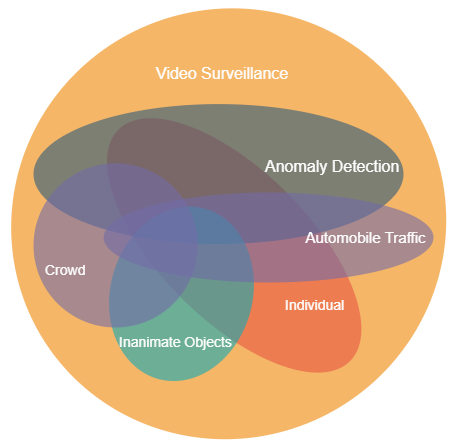}
    \caption{Venn Diagram of Surveillance Targets} 
\end{figure}

\section{Anomaly Detection Methodologies in Video Surveillance}
To improve the safety and well-being of individuals and surrounding, the surveillance has become imperative nowadays. However, it is not possible to view video surveillance scenarios for twenty-four hours and track anomalous events, there is a need for an intelligent surveillance system. 

Anomalies can be contextual, point, or collective. Contextual anomalies are data instances that are considered anomalous when viewed against a certain context associated with the data instance. Point anomalies are single data instances which are different with respect to others \cite{bozcan2021context}. Finally, collective anomalies are data instances which are considered anomalous when viewed with other data instances, concerning entire dataset \cite{cheng2015gaussian}. To detect anomalies in automated surveillance, advanced detection schemes have been developed over a decade. In this survey, we categorize them broadly into; learning-based and modeling-based approaches and further sub-categorize for clear understanding. The anomaly classification based on learning and based on approaches is shown in Figure 4.

\subsection{Learning}
Any event or behaviour that deviates from the normal is called an anomaly. The learning algorithms learn anomalies or normal situations based on the training data which can be labeled or unlabeled. Depending upon the methodologies used the various learning methods for anomaly detection can be classified as;

\begin{enumerate}
    \item Supervised Learning,
    \item Unsupervised Learning and
    \item Semi-supervised Learning.
\end{enumerate}

The anomaly detection classification on the basis of learning is shown in Table 2.
\subsubsection{Supervised learning}

Supervised learning uses labelled dataset to train the algorithm to predict or classify the outcomes. It gives a categorical output or probabilistic output for the different categories. The training data is processed to form different class formulations; single class, two-class or multi-class.\\ 
When the training data contains data samples either of normal situations or anomalous situations only then it is called single class formulation \cite{sodemann2012review}. Since the training data consists of a single class, the labeling is difficult. In a single class approach, if the detector is trained on normal events then the events that fall outside the learned class are classified as anomalous \cite{asad2021multi}. Various approaches to classify and model anomalies with such training data use a 3D Convolutional Neural Network Model \cite{ji20123d}, Stacked Sparse Coding (SSC) \cite{xu2018anomaly}, adaptive iterative hard-thresholding algorithm \cite{zhou2019anomalynet}.\\
Apart from single and two class formulation, an approach where multiple classes of events are learned is called multi-class formulation. In this approach before anomaly detection, certain rules are defined regarding behavior classification. Anomaly detection is then performed using these set of rules \cite{lao2009automatic} \cite{saligrama2012video}. However, this approach has a drawback that, the events that are learned can only be reliably recognized and the events that do not span the learned domain are incorrectly classified. Thus, the multi-class approach may not provide optimum results outside a scripted environment.
%

\begin{figure}
    \centering
    \includegraphics[width=7cm]{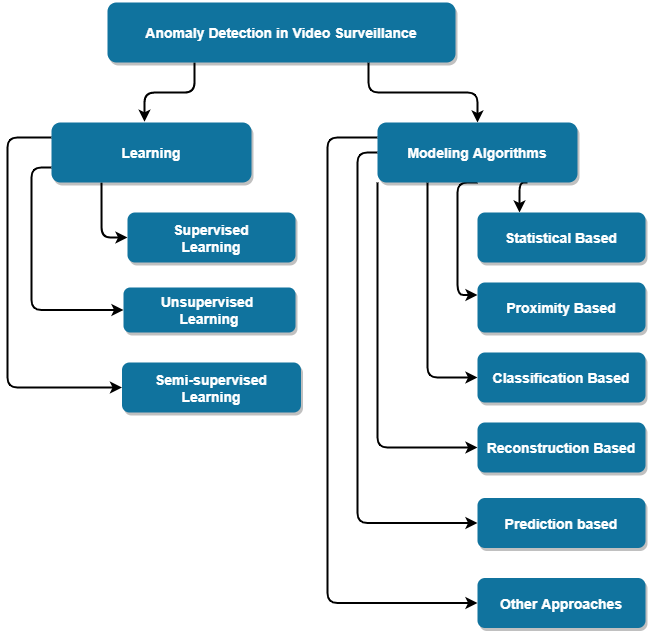}
    \caption{Anomaly Classification} 
\end{figure}

\begin{table}
    \centering
\caption{Anomaly Detection Classification}
\begin{tabular}{l|l}
    Type of Learning & Reference \\
    \hline
    Supervised &    \cite{xu2018anomaly} \cite{zhou2019anomalynet} \cite{lao2009automatic}  \cite{asad2021multi} \cite{ji20123d} \cite{saligrama2012video} \\
    Unsupervised &  \cite{wu2019deep} \cite{nawaratne2019spatiotemporal}  \cite{song2019learning} \cite{kumar2017visual} \cite{lamba2019segmentation} \cite{ren2015unsupervised} \cite{chu2018sparse} \cite{sun2020adversarial} \cite{cocsar2016toward}  \cite{roy2018unsupervised} \cite{perveen2018spontaneous} \cite{chowdhury2020unsupervised} \cite{chriki2020uav} \cite{chriki2021deep} \cite{bansod2020crowd} \\
    Semi-supervised & \cite{tripathi2019convolutional} \cite{wang2019saliencygan} \cite{liu2016semi} \\
\end{tabular}
\end{table}

\subsubsection{Unsupervised Learning}

In unsupervised learning, given a set of unlabeled data, we discover patterns in data by cohesive grouping, association, or frequent occurrence in the data. In this approach, both normal and abnormal training data samples which do not have any label. An algorithm discovers patterns and groups them together with an assumption that the training data consist of mostly normal events and occurs frequently while rare events are termed as anomalous \cite{sodemann2012review}\cite{wu2019deep}. However, because the non-deterministic nature of anomalous events and insufficient training data it is challenging to automatically detect anomalies in surveillance videos. To address these issues, \cite{song2019learning} presented an adversarial attention-based auto-encoder network. It uses a reconstruction error produced by the auto-encoder to diverge abnormal events/anomalies from normal events. 

\cite{kumar2017visual} introduces a trajectory-based hierarchical clustering algorithm framework named Visual Assessment of Tendency (VAT) to detect abnormal trajectories. For anomaly detection in a crowd, \cite{lamba2019segmentation} estimates contour-based trajectory clustering method where the crowd flow is segmented using density-based DBSCAN clustering. For experimentation high density-based UCF, Collective motion, Violent Flows datasets are used to provide optimum results. \\ 

Recently, deep 3-dimensional convolutional network (C3D) \cite{chu2018sparse} is widely used for video surveillance applications. The local spatiotemporal patterns captured by C3D are suitable for video data however, it is tedious to incorporate this supervised model to detect anomalies as there are no categorical labels involved and hence, all the events will be viewed as normal events. To cope with this, self-supervised signals are developed by extracting spatiotemporal patterns in videos and agglomerative clustering is employed to obtain a similarity relationship between the inputs to train C3D. Similarly by utilizing the self-similarity property of the training data, Unsupervised Kernel Learning with Clustering Constraint (CCUKL) is deployed to learn abnormal events where the feature space is learned using Non-negative matrix factorization (NMF) and the clustering degree is measured using support vector data description (SVDD) \cite{ren2015unsupervised}.

Some studies have also proposed to used C3D and adversarial auto-encoder for detecting anomalous events in videos \cite{sun2020adversarial}. The 3D convolution auto-encoder model aims to learn the spatiotemporal patterns and train the auto-encoder by using the de-noising reconstruction error and adversarial learning strategy to detect anomalies without supervision \cite{cocsar2016toward}. To distinguish between new anomalies and normality that evolve, Incremental Spatio-Temporal Learner (ISTL) remain updated about the changing nature of anomalies by utilizing active learning with fuzzy aggregation \cite{nawaratne2019spatiotemporal}. ISTL approach is estimated on a spatiotemporal auto-encoder model which consists of convolution layers and ConvLSTM (Convolution Long-Short Term Memory)layers that learn spatial-temporal regularities. Any anomalies in the scene are periodically monitored by a human observer so that the system dynamically evolves normal behavior using fuzzy aggregation.

For the purpose of action recognition in surveillance scenes \cite{roy2018unsupervised} proposes a Gaussian mixture model called Universal Attribute Modelling (UAM) using unsupervised learning approach. The UAM is also been used for facial expression recognition where it captures the attributes of all expressions \cite{perveen2018spontaneous}

Further, for autonomous vehicles like cars or UAVs (Unmanned Aerial Vehicles) it is very essential to distinguish between normal and anomalous states. Chowdhury et al. estimates the degree of abnormality using an unsupervised heterogeneous system from real-time images and IMU (Inertial Measurement Unit) sensor data in a UAV \cite{chowdhury2020unsupervised}. They also demonstrated a CNN architecture to estimate an angle between a normal image and query image, to provide a measure of anomaly. 
Recently, one-class classifiers \cite{chriki2020uav} \cite{chriki2021deep} are evolved as state-of-the-art for anomaly detection. They use a CNN with One-Class Support Vector Machine (OCSVM) to detect anomalies or abnormalities in the data. 

\cite{bansod2020crowd} uses a Physics theory of momentum which is used to establish a relationship between the foreground region and object motion. By utilizing the statistical and positional features, a histogram of magnitude and momentum (HoMM) is derived, and further by K-means clustering and distance calculation anomalous and non-anomalous events are recognized.

\subsubsection{Semi-supervised Learning}
Semi-supervised learning falls between supervised learning and unsupervised learning. It combines a small amount of labeled data with a large amount of unlabeled data during training. Semi-supervised learning is used where less variety of labeled training dataset is available such as in extracting suspicious events in smart security camera \cite{wang2019saliencygan}. In such situations, Salient Object Detection (SOD) is a commonly used fundamental pre-processing tool for deep learning models like SaliencyGAN (Saliency Generative Adversarial Network). Different combinations of labelled and unlabelled data are used in training of SaliencyGAN to obtain misclassified events. In some applications, Laplacian Support Vector Machine (LapSVM) utilizes unlabeled samples to learn a more accurate classifier \cite{liu2016semi}. It is observed that, there is a considerable improvement in learning accuracy when unlabeled data is used in conjunction with small amount of labelled data. \cite{tripathi2019convolutional} proposes an object tracking method using structural sparse representation where edge detection is deployed to avoid tracking error accumulation in case of occlusion in complex scenes.

\begin{table*}[]
    \centering
    \caption{Categorization of Anomaly Detection Techniques}
    \begin{tabular}{l|l|l|l}
       
       Approach  & Ref & Technique & Highlights \\
        \hline
         
        Statistical & \cite{cheng2015gaussian} & Gaussian Process  & Hierarchical Feature Representation, Gaussian Process  \\
         Based & & Regression &  Regression. STIP ( spatio-temporal interest points) is used   \\
        & & & to detect local and global video anomaly detection;  \\
        &&& Dataset: Subway, UCSD, Behave, QMUL Junction;\\
        & & & Parameters: AUC, EER\\
        &&& (AUC: Area Under Curve), EER: Equal Error Rate)\\
        
        & \cite{kaltsa2015swarm} & Histogram Based & HOG, Histograms of Oriented Swarms (HOS) KLT interest \\
     & & Model  & point tracking are used to detect anomalous event in crowd;  \\ 
     & & & Aims at achieving high accuracy and low computational cost; \\
     &&& Dataset used: UCSD, UMN; Parameters: ROI, EER, DR; \\
     &&& (ROI: Region of Interest, DR: Detection Rate)\\
     
     & \cite{bansod2020crowd} & Histogram Based   & Crowd anomaly detection using HoMM (histogram of \\
     & &   &  magnitude and momentum); background removal,\\
     & &  & feature extraction (optical flow), anomaly detection \\
     & & & (using K-means clustering); Dataset: UCSD, UMN;\\
     & & &  Parameters: AUC, EER, RD (Rate  Reduction) \\
& & &  True Positive Rate (TPR), False Positive Rate (FPR)\\
     
      & \cite{nguyen2015bayesian} & Bayesian Model & Bayesian Non-Parametric (BNP) approach,  Hidden Markov\\
     & & & Model (HMM) and Bayesian non-parametric factor  \\
     & & & analysis is employed for data segmentation and pattern\\
     & & &  discovery of abnormal events. Dataset: MIT\\
     & & & Parameters: Energy Distribution \\

     & \cite{isupova2016anomaly} & Bayesian Model & Bayesian non-parametric dynamic topic model is used. \\ 
     & & &Hierarchical Dirichlet process (HDP) is used to detect anomaly.\\ 
     & & & Dataset: QMUL-junction; Paremeters: ROC, AUC\\
     &&& (ROC: Receiver Operating Characteristics)\\

       & \cite{sabokrou2018deep} & Gaussian Classifier & Fully convolutional neural network, Gaussian classifier is \\
    &  &  &  used. It extracts distinctive features of video regions to  \\
     &  &  & detect anomaly. Dataset: UCSD, Subway; \\
     &&&  Parameters: ROC, EER, AUC; \\
     & & & Aims to increase the speed and accuracy.\\

     & \cite{yuan2014online} & Histogram Based & Structural Context Descriptor (SCD), SHOF, 3D DCT  \\
     & & Model & Selective Histogram of Optical Flow (SHOF),\\
     & & & Discrete Cosine Transform (DCT)\\
     & & &  object tracker, Spatiotemporal analysis for abnormal   \\
     & & & detection is crowd using Energy Function  \\
     & & & Dataset: UMN and UCSD; \\
     &&& Parameters: ROC, AUC, TPR, FPR\\

     \hline
     Proximity  & \cite{colque2016histograms} & Histogram  &  Histogram of optical flow and motion entropy (HOFME) is \\
     Based & & &  used to detect the pixel level features diverse anomalous  \\
     & & & events in crowd  anomaly scenarios as compared with  \\
     & & &  conventional features. Nearest Neighbor threshold is used  \\
     &&&  by HOFME.\\
     &&& Dataset: UCSD, Subway, Badminton;  \\
     &&& Parameters: AUC, EER \\
     
     &  \cite{liu2018accumulated} & Accumulated & ARD method is used for large-scale traffic data and detect\\
     && Relative &   outliers;  Dataset: Self-deployed  \\
     & & Density (ARD) & Parameter: Detection Success Rate (DSR) \\
     &\cite{hu2018anomaly} & Density Based & A weighted neighborhood density estimation is used to detect  \\
     & & & anomalies. Hierarchical context-based local kernel regression. \\
     & & & Dataset: KDD, Shuttle; Parameters: Precision, recall \\

\hline
  \end{tabular}
    
    \label{table_2}
\end{table*}

\begin{table*}[]
    \centering
    \begin{tabular}{l|l|l|l}
       
       Approach  & Ref & Technique & Highlights \\
       \hline
       
       Proximity & \cite{lamba2019segmentation} & Trajectory Extraction & Foreground segmentation by using active contouring \\
     Based & & Flow analysis & Density based DBSCAN Clustering.\\
     && &Dataset: UCF Web, Collective Motion, and Violent Flows \\
     & & & Parameters: MAE (Mean Absolute Error) and F-score \\
     
     \hline
       
      Classification & \cite{mo2013adaptive} & Adaptive Sparse  & Trajectory-based video anomaly detection using joint  \\
 Based & & Representation & sparsity model. Dataset: CAVIAR. Parameters: ROC\\
 
  & \cite{wang2014detection}  & One-Class  & OCC based anomaly detection techniques using SVM. \\
   & & classification & Pixel level features are extracted. \\
     &&&  Dataset: PETS2009 and UMN; Parameters: ROC \\
     
      & \cite{nikouei2018smart} & Harr-Cascade,  & Smart Surveillance as an Edge Network using Harr-\\
     & & HOG feature  & Cascade, SVM, L-CNN. Provides fast object detection  \\
     & & extraction & and tracking. \\
     & & & Dataset: VOC07, ImageNet; Parameters: Accuracy\\

 & \cite{xu2018real} & Stacked Sparse & Intraframe classification strategy; Dataset: UCSD, Avenue, \\
  & & Coding  & Subway; Parameters: EER, AUC, Accuracy\\

 & \cite{shi2017sequential} & Sequential Deep   & Dense trajectories are projected into 2D plane, \\
 &  & Trajectory &  long term motion CNN-RNN network is employed.\\
& &  Descriptor & Dataset: KTH, HMDB51, UCF101; Parameters: Accuracy\\

&\cite{ghosh2020edge} & Autoencoders,  & Feature learning with deep learning, autoencoder is placed \\
 & &Data Reduction & on the edge, decoder part is placed on the cloud\\
& & & Dataset: HAR, MHEALTH; Parameters: Accuracy \\

& \cite{ke2016real} & Kanade–Lucas–& k-means clustering, connected graphs and traffic flow\\
&  & Tomasi  (KLT)& theory are used estimate real-time parameters from \\
& &  tracker  & aerial videos; Dataset: Manual;  \\
&&& Parameters: speed, density, volume of vehicles\\

 \hline

       Reconstruction & \cite{lei2020discriminative} &  Hyperspectral Image  & Discriminative reconstruction method for HSI anomaly  \\
& & (HSI) Analysis &  detection with spectral learning (SLDR). Loss Function  \\
& & & of SLDR Model is generated. \\
& & & Dataset: ABU,  San Diego\\
& & & Parameters: ROC, AUC\\

& \cite{xu2019low} & Hyperspectral  & Low-rank sparse matrix decomposition (LRaSMD) \\
& & Imagery (HSI) & based dictionary reconstruction is used for anomaly \\
& & & detection. Dataset: STONE, AVIRIS\\
& & & Parameters: ROC \\

& \cite{chu2018sparse} &  3D convolutional  & C3D network is used to perform feature extraction \\
& & network  &  and detect anomaly using sparse coding, DL. \\
& & & Dataset: Avenue, Subway, UCSD ;\\
&&&  Parameters: AUC, EER\\

& \cite{wu2019deep} & Deep OC Neural  & One stage
model is used to learn compact features   \\
& & Network & and train a DeepOC (Deep One Class)classifier. \\
& & & Dataset: UCSD, Avenue, Live Video; Parameters: ROC\\ 

& \cite{jiang2020discriminative}& Generative  & Anomaly detection using Generative Adversarial
\\
& & Adversarial & Network for hyperspectral images (HADGAN).\\
& &  Network (GAN) & Dataset: ABU, San Diego, HYDICE\\
& & & Parameters: ROC, Computing time\\

& \cite{song2019learning} & Adversarial  & Normal patterns are learnt through adversarial \\
& &  attention based , & attention based auto-encoder and anomaly is detected. \\
& &  auto-encoder GAN & Dataset: ShanghaiTech, Avenue, UCSD, Subway;\\
&&&  Parameters: AUC EER \\

& \cite{sun2020adversarial} & Adversarial 3D Conv,  & spatio temporal patterns are learnt using\\
& & Autoencoder & adversarial 3D Conv, Autoencoder to detect abnormal \\
& & & events in videos; Dataset: Subway, UCSD, \\
&&&  Avenue, ShanghaiTech; Parameters: AUC/EER\\

 \hline

  \end{tabular}
    
    \label{table_3}
\end{table*}

\begin{table*}[]
    \centering
    \begin{tabular}{l|l|l|l}
       
       Approach  & Ref & Technique & Highlights \\
        \hline
  Reconstruction   & \cite{yu2016content} & Sparse  & Sparsity based reconstruction  method is used with low  \\
Based& & Reconstruction  & rank property to  determine abnormal events.   \\ 
& & & Datasets: UCSD, Avenue; Parameters: ROC, AUC, EER\\

 & \cite{zhang2015abnormal} & Sparsity based  & Abnormal event detection in traffic surveillance using  \\
 & & method & low-rank sparse representation (CLSR).\\
 & & & Dataset: UCSD, Subway, Avenue; \\
 &&&  Parameters: AUC, EER\\

\hline   
 Prediction & \cite{liu2018future} & Video prediction & Spatial/motion constraints are used for future  \\

 Based & & framework & frame prediction for normal events and \\
 & & &identifies abnormal events; Dataset: CUHK, UCSD, \\
 & & & ShanghaiTech; Parameters: AUC \\
 
 & \cite{nawaratne2019spatiotemporal} & Incremental Spatio- & ISTL, an unsupervised deep learning approach with  \\
 && Temporal Learner &  fuzzy aggregation is used to distinguish between \\
&&&  anomalies that evolve over time in assistance with  \\
&&& spatiotemporal autoencoder, ConvLSTM.  \\
&&& Dataset: CUHK, Avenue, UCSD;  Parameters: AUC, EER\\
 
         & \cite{lai2019lstm} & LSTM, Cross  & Recognizing industrial
equipment in manufacturing  \\
 & & Entropy & system using edge computing; Big Data, smart meter\\

& & & Dataset: Manual ; Parameter: Accuracy\\

& \cite{cheng2020securead} & Optical flow, GMM, & Detection accuracy, less computational time \\ 

& & HoF  & Dataset: UMN, UCSD, Subway, LV; \\
&&& Parameter: AUC, EER;\\

       & \cite{zhou2019anomalynet} & spatiotemporal feature  & CNN, Adaptive
ISTA and SLSTM \\
 & & extraction & Dataset: CUHK Avenue, UCSD, UMN ;  \\ 
&&& Parameter:AUC EER\\

\hline

 Other  & \cite{li2018road} & Fuzzy theory & Anomaly detection in road traffic using Fuzzy  \\
 
Approaches & & & theory. The Gaussian distribution model is trained.\\
 & & & Dataset: SNA2014-Nomal\\
 & & & Parameter: Accuracy, False Detection Rate \\
 
      &  \cite{mo2013adaptive} & Sparse Reconstruction &  Joint sparsity model for abnormality detection, \\
      & & & multi-object anomaly detection for real world \\
      & & & scenarios. \\
      & & & Dataset: CAVIAR \\
      & & & Parameters:ROC\\

     & \cite{li2020adaptive} & Sparsity Based & Video surveillance of traffic. Background subtraction.\\
     & & &  Non-convex optimization; Generalized Shrinkage\\
     & & &  Thresholding Operator (GSTO), Joint estimation\\
     & & & Dataset: I2R, CDnet2014 \\
     & & & Parameters: F-measure \\
     
     &  \cite{guo2019detecting} & Frequency Domain & Detection of Vehicle Anomaly using Edge Computing\\
     & & (Power Spectral & high-frequency correlation, sensors, reduces \\
     & &  Density; Correlation) &  computation overhead, privacy\\
     & & & Dataset: Open Source Platform \\
     & & & Parameters: FPR, TPR, ROC\\

\hline

  \end{tabular}
    
    \label{table_4}
\end{table*}

\subsection{Modeling Algorithms for Anomaly Detection}
\subsubsection{Statistical Based}

In statistical based approach the parameters of the model are learnt to estimate anomalous activity. The aim is to model the distribution of normal-activity data. The expected outcome under the probabilistic model will have higher likelihood for normal-activities and lower likelihood for abnormal activities \cite{bergman2020classification}. Statistical approaches can further be classified as parametric method and non-parametric method. Parametric methods assume that the normal-activity data can be represented by some kind of probability density function \cite{kumaran2019anomaly}. Some methods use Gaussian Mixture Model (GMM) which works only if the data satisfies the probabilistic assumptions implicated by the model \cite{cheng2015gaussian}. Non-parametric statistical model is determined dynamically from the data. Examples of non-parametric models are histogram-based \cite{kaltsa2015swarm} models, HoMM (histogram of magnitude and momentum) based models \cite{bansod2020crowd}, Bayesian \cite{nguyen2015bayesian} \cite{isupova2016anomaly} models. Recently, efficient way to detect and localize anomalies in surveillance videos is to use Fully Convolutional Networks (FCNs) \cite{sabokrou2018deep} and Structural Context Descriptor (SCD) \cite{yuan2014online}.

The detailed categorization of anomaly detection techniques are shown in Table 3.

\subsubsection{Proximity Based}
When the video frame is sparsely crowded it is easier to detect anomalies, but it becomes a tedious job to find irregularities in densely crowded frame. Proximity based technique utilizes the distance between the object and its surrounding to detect anomalies. In \cite{colque2016histograms}, a distance-based approach is used that assumes normal data has dense neighborhood and anomalies are identified by their proximity to their neighbours. Further, density-based approaches identify distinctive groups or clusters depending upon the density and sparsity and the anomaly is detected \cite{liu2018accumulated} \cite{hu2018anomaly} \cite{lamba2019segmentation}.

\subsubsection{Classification Based} 
Another commonly used methods of anomaly detection are classification based which aims to distinguish between events by determining the margin of separation. In \cite{mo2013adaptive}, Support Vector Machine (SVM) uses classic kernel to learn a feature space with to detect anomaly. Further, a non-linear one-class SVM trained with histogram of optical flow orientation to encode the moving information of each video frame \cite{wang2014detection}. Aiming at intelligent human object surveillance scheme, Harr-cascade and HOG+SVM is applied together to enable a real-time human-objects identification \cite{nikouei2018smart}. Some approaches utilize object trajectories to understand the nature of object in the scene and detect anomalies, various tracking algorithms \cite{xu2018real}\cite{parate2017structurally}\cite{parate2018global} are used to estimate trajectories of an object. Trajectory based descriptors are also widely used to capture long term motion information and to estimate the dynamic information of foreground objects for action recognition \cite{shi2017sequential}. Some approaches use PCA \cite{ghosh2020edge}, K-means \cite{ke2016real} for detection of anomaly.

\subsubsection{Reconstruction Based}
In reconstruction-based techniques, the anomalies are estimated based on reconstruction error. In this technique every normal sample is reconstructed accurately using a limited set of basis functions whereas abnormal data is observed to have larger reconstruction loss \cite{kumaran2019anomaly}. Depending on the model type, different loss functions and basis functions are used. Some of the methods use Hyperspectral Image (HSI) \cite{lei2020discriminative}, \cite{xu2019low}, 3D convolution network \cite{chu2018sparse}.

Recently, a deep neural network DeepOC in \cite{wu2019deep} can simultaneously train a classifier and learn compact feature representations. This framework uses the reconstruction error between the ground truth and predicted future frame to detect anomalous events. Another set of methods use Generative Adversarial Network (GAN) \cite{kumar2020generative} to learn the reconstruction loss function \cite{jiang2020discriminative}. GAN based auto-encoder proposed in \cite{song2019learning} produce reconstruction error and detect anomalous events by distinguishing them from the normal events. Further, an adversarial learning strategy and denoising reconstruction error are used to train a 3D convolutional  auto-encoder to discriminate abnormal events \cite{sun2020adversarial}.

Another paradigm to detect anomalous events is by exploiting the low-rank property of video sequences. In the phase of learning, lower construction costs are assigned to instances that describe important characteristics of the normal behavior. Depending on low-rank approximation, a weighted sparse reconstruction method is estimated to describe the abnormality of testing samples \cite{yu2016content} \cite{zhang2015abnormal}.

\subsubsection{Prediction based}
Prediction-based approach use known results to train a model. Such a model predicts the probability of the target variable based on the estimated significance from the set of input variables. In prediction-based approach, the difference between the actual and predicted spatio-temporal characteristics of the feature descriptor is calculated to detect the anomaly \cite{liu2018future}. Also, Incremental Spatio-Temporal Learning (ISTL) approach with fuzzy aggregation is used to distinguish anomalies that evolve over time \cite{nawaratne2019spatiotemporal}.
Further, in sequence prediction, Long Short Term Memory (LSTMs) are very powerful as they store past information to estimate future predictions . LSTM networks are used to learn the temporal representation to remember the history of the motion information to achieve better predictions \cite{lai2019lstm}. To enhance the approach, \cite{cheng2020securead}  integrates autoencoder and LSTM in a convolutional framework to detect video anomaly. Another technique of learning spatiotemporal characteristics is estimating an adaptive iterative hard-thresholding algorithm (ISTA) where a recurrent neural network is used to learn sparse representation and dictionary to detect anomalies \cite{zhou2019anomalynet}. 

\subsubsection{Other Approaches}

To handle complex issues in traffic surveillance, \cite{li2018road} estimates a fuzzy theory and propose a traffic anomaly detection algorithm. To perform the state evaluation, virtual detection lines are used to design the fuzzy traffic flow, pixel statistics are used to design fuzzy traffic density, and vehicle trajectory is used to design the fuzzy motion of the target. To recognize abnormal events in traffic such as accidents, unsafe driving behavior, on-street crime, traffic violations, \cite{mo2013adaptive} proposes adaptive sparsity model to detect such anomalous events. Similarly, \cite{li2020adaptive} estimates sparsity based background subtraction method and shrinkage operators. Other approaches also include \cite{guo2019detecting}, which uses high-frequency correlation sensors to detect vehicle anomaly.

\section{Edge Computing}

 In traditional video surveillance systems, raw video data gathered from all visual sensors is sent to the centralized servers for storage and further processing. Some approaches use cloud-based computing methodologies for the same. However, the inevitable network latency and bandwidth requirements are not suitable for real-time applications, especially in time-critical applications such as anomaly detection. An edge computing \cite{shi2016edge} is proposed as a potential solution to this problem as it requires low bandwidth and is applicable where network latency and privacy are concerns \cite{varghese2016challenges}.
 
 Edge computing is a distributed, decentralized computing method that provides on-board computation and storage of data. Most of the data produced at the device are processed at the device itself \cite{schneible2017anomaly}. With the advancement in the terminal or edge devices, few contributions are observed in detecting anomalies at the edge or terminal devices. Schneible et al. present a federated learning approach in which autoencoders are deployed on edge devices to identify anomalies. Utilizing a centralized server as a back-end processing system, the local models are updated and redistributed to the edge devices \cite{8170817}. Despite the rapid development of learning methods, CNNs, and edge devices for computer vision purposes particularly, the gap between software and hardware implementations is already considerable \cite{8114708}.
 
 \begin{figure}
    \centering
   \includegraphics[width=7cm]{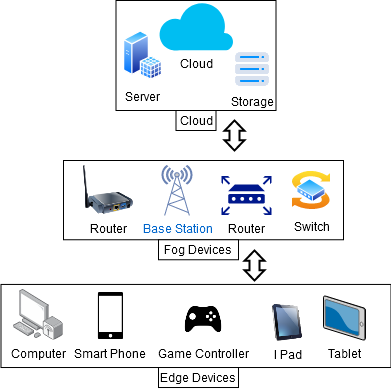}\\
    \caption{Architectural Overview of Edge Computing}
\end{figure}

The general architectural overview of the edge computing paradigm is shown in Figure 5. The top-level entities are cloud storage and computing devices which comprise data centers and servers. The middle level represents fog computing. Any device with compute capability, memory, and network connectivity is called a fog node. Examples of fog devices are switches, routers, servers, controllers. The bottom-most part of the pyramid includes Edge Devices like sensors, actuators, smartphones, mobile phones. These terminal devices participate in processing a particular task using a user access encryption \cite{kang2017privacy}. 

\section{Anomaly Detection using Edge Devices}

The terminal processing at edge devices for automated surveillance is considered to be the alternative for cloud and network-based processing when response time is a relatively important parameter. Over the decade, some approaches have been crafted for automated surveillance/object tracking using edge devices and a few of them talk about anomaly detection in video surveillance as summarized in Table 4.

Traditional computer vision methods such as feature-based classification approaches are noticeable candidates for edge application; for example, Harr-Cascaded and HOG+SVM algorithms are widely used for human detection in surveillance. Nikouei et al. \cite{nikouei2018smart} deployed an SVM classifier trained on Harr-Cascade and HOG feature at the edge and introduces a Lightweight Convolutional Neural Network (L-CNN) for smart surveillance. The model is trained using VOC07 and ImageNet datasets with the MXNet platform for neural networks. Again, Histogram of Oriented Gradients (HOG) and Support Vector Machine (SVM) along with a Kernelized Correlation Filters (KCF) is implemented to be deployed on Raspberry pi 3 which is an only CPU edge device \cite{xu2018real}. On similar lines, a Kerman algorithm \cite{nikouei2019toward}, which is a combination of Kernelized Kalman filter, Kalman Filter (KF), and Background subtraction (BS) is proposed to achieve enhanced performance on edge. Kernelized Kalman filter is based on decision trees and is suitable for human object tracking. Kerman outperforms Haar-cascade, HOG+SVM, SSD Google Net in terms of FPR (False Positive Rate), FPS (Frames Per Second), and speed and can track/re-find the human objects in real-time.\\

The traditional methods, though performing well in some scenarios are not as accurate as of the modern learning approaches. Wang et al. propose SaliencyGAN  \cite{wang2019saliencygan}, that uses a semi-supervised method for Salient Object Detection in the fog-IoT network. Salient Object Detection (SOD) is a useful tool for human-object detection and tracking. However, due to diverse fog devices, implementing SOD on fog nodes is a cumbersome task.  The proposed SaliencyGAN is trained with 10\% to 100\% labeled training data. SaliencyGAN showcased comparable performance to the supervised approaches when the labeled data reached 30\% and outperformed the unsupervised and weakly supervised approaches. Unlike, SaliencyGAN, Zhao et al. opt lightweight deep learning-based model to reduce network occupancy and reduce the system response delay by using edge and cloud computing together \cite{zhao2020lightweight}. Altogether the system has benefits of low time response and resource management.
Recently, to ensure passenger safety in public transportation, Ajay et al. propose a Binary Neural Network (BNN) based on real-time emotion detection of passengers using edge computing devices \cite{ajay2021binary}. Here the facial expressions are classified into six categories namely: fear, happy, sad, angry, surprise, and disgust. The LBP-BNN based improved and faster emotion recognition model keeps the track of facial expressions of people and is suitable for is used for applications including tracking of facial expressions for physically disordered people in hospitals.

\begin{table*}
\centering
 \caption{Anomaly Detection at the Edge}
 \begin{tabular}{l l l l l l}
Ref & Features & Learning & Anomaly Criteria & Dataset & Parameters\\
\hline

\cite{ajay2021binary}  & Accuracy & Binary Neural & Recognition & JAFFE dataset & Accuracy\\
&&Network; FPGA &of facial &&\\
&&system design&emotions&&\\
\hline

 \cite{deng2021air} & Reliability, & Air-Ground & Target tracking  & Self deployed & Probability of   \\
 &timeliness, &  Surveillance & in visually & &target retrieval,\\
 & opportunistic & Sensor Network &  obstacle blocking & &delay, workload \\
 &&&scenarios.\\

\hline
 \cite{zhao2020lightweight} & Reduce network   & Intelligent Edge  &  Cloud,  & Self deployed& Accuracy \\

         & occupancy and &      Surveillance   &  DL, Edge     &  & Model loss  \\
        &  system response  & & & \\
        & delay &&&\\
        
        \hline
       \cite{ke2020smart} & Smart parking & CNN, HOG  & Parking & Self deployed & Accuracy\\
       &  &      &    Surveillance    &   \\

       \hline
   \cite{nikouei2019toward}  & Real-time, good & LCNN, Kerman & Tracking  &  VOC07, VOC12 & Accuracy \\
         & accuracy, with  & (KCF, KF, BS) & Human Objects        &  & Precision \\
         & limited resources & & & \\
         \hline
   \cite{wang2019saliencygan}  & Increases & SaliencyGAN & Object & 
  & MAE    \\
        & computing & Deep SOD CNN &   Detection &  PASCALS  & F-Measure \\
         & performance & Adversarial  & /Tracking  & & Precision  \\
          & & Learning  & &  &Recall \\
          && Semi-supervised && & \\
         
         \hline
         
          \cite{muhammad2019edge}  & Processing   & LCNN & Smoke   & ImageNet & FP, FN, 
 \\
       & requires &  &  Detection    &  &  Accuracy\\
      & less memory & &in Foggy&& precision, \\
      &&&Surveillance &&  recall,  \\
      &&&&& F-measure\\
      \hline
      
       \cite{lai2019lstm} & Parallel & LSTM, & Industrial   & Self deployed & Accuracy\\
      & Computing to  &     Cross Entropy &  Electrical  &     &    \\
       &improve&  &Equipment&& \\
       & efficiency &  &&& \\
    \hline 
    
    \cite{guo2019detecting} & Reduced & high-frequency  & Vehicle  & Open Source & FPR, TPR, \\
       &  computation , &    correlation, &    Anomaly    & Platform & ROC   \\
       & overhead & sensors&&& \\
       & privacy  &&&& \\
       
       \hline
       
       \cite{chen2019distributed} & Balances  & DIVS & Vehicle  & Self deployed & Efficiency  \\
       & computational &       &   classification   &   &  \\
      & power and &&& \\
      & workload &&&& \\
      \hline

   \cite{xu2018real} &  Low   & HOG,SVM, & Human Object  & Self deployed   & Speed,\\
      & computational  &   KCF &  Tracking  &  & Performance   \\
     & cost with & & &   & \\
     &  high accuracy  & & &  & \\
     & and performance & &  \\
     \hline

      \cite{li2018deep}  & Improved  & CNN, DL & Industry  & Self deployed & FPR, TPR,\\
       & efficiency  &              & Manufacture  &   &  ROC   \\
       & & &  Inspection  &\\
       \hline

   \cite{wang2018offloading} & Resource  & FORT & Traffic & Self deployed & average  \\
      & management,   &     (Real-Time Traffic  & Management   &      & response    \\
    & reduction in& management    &System& & time \\
    & response time  &in Fog)&&& \\

       \hline
       
\end{tabular}\\
\end{table*}

\begin{table*}
\centering
 \begin{tabular}{l l l l l l}       
       
        \hline
     
      \cite{nikouei2018smart} & Processing   & Harr-Cascade,  & Smart  & ImageNet & FP, FN, Accuracy\\
        & requires &  SVM, L-CNN  &   Surveillance         &  &   precision, recall,  \\
       &less memory&&&& F-measure (F) \\
       
       \hline
       
       \cite{kang2017privacy}  & Reduce  & Privacy  & Smart IoV & Self & Entropy comparison  \\
        &  computational  &   -Preserved    &  (Internet of Vehicles)        & deployed&  Nash Equilibrium, \\
       &  overheads and   & Pseudonym  & address privacy &&   \\
       & enhance location & Algorithm  &location issues &&   \\
       & privacy &&&&\\

       \hline
       
\end{tabular}\\
\end{table*}

Further, UAV (Unmanned Aerial Vehicle) proves to be beneficial as it offers good performance in sight-blocking scenarios \cite{chriki2021deep}. On one hand there is a cloud-based tracking system where \cite{zhai2020smart} proposes a cloud-enabled autopilot drone system in video surveillance that uses a Deep Neural Network (DNN) for abnormal event detection. The problem with such a kind of implementation is that the time response is large as the amount of the data to be transferred is large. As this process needs to be done in real-time it becomes tedious to locate the target. To address this challenge, Air-Ground Surveillance Sensor Network (AGSSN) tracking system \cite{deng2021air} is proposed which is based on edge computing. To reduce latency and huge network communication overhead, an option of dividing the computationally expensive tasks among the edge nodes is always open and termed as task offloading \cite{wang2018offloading}.\\

To detect anomalies in traffic surveillance, Chen et al. deployed a Distributed Intelligent Video Surveillance (DIVS) system \cite{chen2019distributed} on an edge computing environment. It includes multi-layer edge computing architecture and a distributed Deep Learning (DL) training model. To reduce network overhead and gain workload balance a multi-layer edge computing architecture is employed for vehicle classification and traffic flow prediction. The experimental setup includes 200 monitoring terminals and 35 EC servers to monitor traffic for 7 days. The results show that the execution time is less even if we increase the number of nodes or number of tasks. An Edge Computing-based Vehicle Anomaly Detection (EVAD) scheme is proposed in \cite{guo2019detecting} to prevent attacks on vehicles. It detects anomalies in the intra-vehicle system by sensors using edge computing. To reduce the computation overhead and improve security, the correlations are organized in the form of a ring architecture. Further, an attempt to identify a real-time parking occupancy is made in \cite{ke2020smart}. This Angle Lake parking garage experiment employed edge devices empowered with a single shot multi-box detector  (SSD-Mobilenet) and is implemented using Tensorflow Lite. This experiment aims to track multiple objects for vehicle parking and occupancy judgments under different environmental conditions such as; rain, fog, sunlight, snow, weekend, weekday, day, and night. The data transmission volume is kept small to be handled by the limited network bandwidth. Results show the network latency will always exist and increases with an increase in the number of cameras employed for the surveillance.\\

Moreover, the edge is also employed to detect industrial anomalies, for example, Muhammad et al. proposed a real-time CNN-based smoke detection system surveillance system for foggy environment \cite{muhammad2019edge}. Owing to the problems related to foggy environmental conditions the author focuses on building a lightweight CNN model on MobileNet V2 and tested different conditions like; smoke, without smoke, smoke with fog, and smoke without fog. In industries, to detect manufacturing anomalies a ”DeepIn” model is proposed in \cite{li2018deep}. It is composed of three modules (fog side,  back-end communication module, server-side) designed using CNN layers. Fog side computing module is used for computational offloading, back-end communication module is used for data exchanges and command traffic, and finally, the server-side is used for defect classification and degree regression.\\

Although many intelligent surveillance methods based on machine learning algorithms are available, it is still challenging to efficiently migrate those smart algorithms to the edge due to the very restricted constraints on resources. However, attempts are made to combine the edge computing architecture with the parallel computing of an artificial neural network \cite{lai2019lstm}. Also, an Edge Artificial Intelligence seems to be a promising technology that combines edge computing, artificial intelligence, and Internet-of-Things (IoT) that migrates computation workloads from central cloud to the edge of the network but has its challenges; security and accessibility \cite{kang2017privacy}, the balance of workload among the edge nodes under the complicated scenarios \cite{wang2018offloading}, synchronization of distributed models in an edge computed environment \cite{chen2019distributed}, and reduced network occupancy and reduced system response \cite{zhao2020lightweight}.

\subsection{Datasets}
There are many publicly available data-sets for validating the surveillance and anomaly detection algorithms. UCSD \cite{mahadevan2010anomaly}, CUHK \cite{nawaratne2019spatiotemporal}, Avenue \cite{lu2013abnormal}, UMN \cite{sabokrou2017deep} and Subway \cite{adam2008robust} are some of the popularly used data-sets for anomaly detection consisting of individuals, crowd, objects, vehicles and human object interaction. A detailed list of datasets corresponding to their surveillance targets is given in Table 5.

Other datasets that are often found in the literature are Shanghai \cite{luo2017revisit}, Badminton \cite{colque2016histograms} Behave and QMUL Junction \cite{cheng2015gaussian}, Mind's Eye and Vanaheim dataset \cite{cocsar2016toward}. These datasets include normal videos and abnormal videos for training and testing purposes depending upon the application. For example, the normal events in the CUHK dataset include pedestrians walking on the footpath, group of pedestrians congregating on the footpath whereas abnormal events include individuals littering, walking on the grass, walking towards the camera, and carrying suspicious objects \cite{nawaratne2019spatiotemporal}.

Events in the UCSD dataset include events captured from various crowd scenes that range from sparse to dense. The data-set represents different situations like; walking on the road, walking on the grass, vehicular movement on the footpath, unexpected behavior like skateboarding, etc. \cite{nawaratne2019spatiotemporal}.

Avenue anomalous dataset includes a random person running, any abandoned object, person walking with an object \cite{zhou2019anomalynet}. UVSD dataset includes individuals and vehicles while DAVIS dataset is composed of various objects (human, vehicles, animals) to obtain the class diversity \cite{zhang2020industrial}. 
Anomalous situations in the Subway data-set include walking in the wrong way (people entering the exit gate) and jumping over the ticket gate \cite{colque2016histograms}. Uturn dataset is a video of a road crossing with trams, vehicles, and pedestrians in the scene. The abnormal activity videos cover illegal U-turns and trams \cite{saligrama2012video}.
Vanaheim Dataset consists of videos containing people passing turnstiles while entering/exiting stations recorded in metro stations \cite{cocsar2016toward}. The abnormal events encountered were a person loitering, a group of people suddenly stopping, a person jumping over turnstiles.

Some authors have also used live videos for the implementation of their respective methods \cite{wu2019deep}. Anomalous events from live videos like an accident, kidnapping, robbery, crime (a man being murdered) are seen in the literature. 

To evaluate the anomaly detection model, Mini-Drone Video Dataset (MDVD) has been shot in a car parking area\cite{chriki2020uav}\cite{chriki2021deep}. Anomalous behavior in MDVD includes people fighting, wrongly parked vehicles, or people stealing items or vehicles.

Various algorithms have been developed to tackle challenges in video surveillance in different datasets.

\begin{table}
\centering
 \caption{List of Datasets}
\begin{tabular}{l|l|l}

   Dataset  & References & Surveillance Target \\
   \hline
  
   UCSD &  \cite{wu2019deep} \cite{xu2018anomaly} \cite{singh2017graph} \cite{zhou2019anomalynet} \cite{yuan2014online} &  Automobile, Individual, Crowd (Public Places)\\
   & \cite{colque2016histograms} \cite{sabokrou2017deep}  \cite{sabokrou2018deep} \cite{nawaratne2019spatiotemporal}  \cite{cheng2015gaussian} \\
   &  \cite{saligrama2012video}  \cite{song2019learning}  \cite{chu2018sparse}  \cite{sun2020adversarial}  \cite{yu2016content} \\
   &  \cite{liu2018future} \\

   Avenue, CUHK & \cite{wu2019deep} \cite{xu2018anomaly}  \cite{zhou2019anomalynet} \cite{nawaratne2019spatiotemporal}  \cite{song2019learning} & Individual, Crowd, Object  (Public Places)  \\
   & \cite{chu2018sparse}  \cite{sun2020adversarial} \cite{cheng2020securead}  \cite{liu2018future}   \\

   UMN & \cite{singh2017graph} \cite{zhou2019anomalynet}  \cite{sabokrou2017deep}  \cite{saligrama2012video} \cite{yu2016content} & Individual, Crowd (Public Places) \\

   Subway & \cite{wu2019deep} \cite{xu2018anomaly} \cite{kaltsa2015swarm} \cite{colque2016histograms} \cite{sabokrou2018deep}    & Individual, Crowd (Entrance and Exit of Subway \\
   & \cite{cheng2015gaussian} \cite{saligrama2012video}  \cite{song2019learning} \cite{chu2018sparse}  \cite{sun2020adversarial}   & Stations)\\ 
   & \cite{cocsar2016toward} \\

   Uturn & \cite{saligrama2012video} & Objects (Vehicles), Individuals (Pedestrians)\\
   Vanaheim & \cite{cocsar2016toward} & Individual, Crowd (Metro Stations)\\
   Mind's Eye & \cite{cocsar2016toward} & Individual, Crowd, Objects (Vehicles) (Parking Area)\\
   UVSD, DAVIS  & \cite{zhang2020industrial} & Individuals, Objects (Vehicles)\\
    Shanghai & \cite{song2019learning} \cite{sun2020adversarial} \cite{liu2018future} \cite{luo2017revisit}  & Individual, Crowd, Objects (Vehicles) (Public Places)\\
    Badminton & \cite{colque2016histograms} & Individual, Crowd (Badminton Game)\\
    Behave &  \cite{cheng2015gaussian} & Individual, Crowd Public Places)\\
    ABU Dataset & \cite{lei2020discriminative} \cite{jiang2020discriminative} & Objects (Aerial View)\\
    MDVD & \cite{chriki2020uav}\cite{chriki2021deep} & Objects (Aerial View)\\
    IEEE SP Cup-2020 & \cite{chowdhury2020unsupervised}  \cite{chowdhury2020anomaly} & Objects (Lobby, Parking Area)\\
    AU-AIR & \cite{bozcan2021context} & Objects (Aerial View)\\
    Facial Expression  & \cite{ajay2021binary} & Individuals (Facial Expressions) \\
    2013 (FER-2013) & & \\
    ISLD-A & \cite{angelini2019privacy} & Individuals, Objects (Human Action Recognition)\\

    Self-deployed &  \cite{lao2009automatic}  \cite{ke2016real} \cite{nieto2018automatic}  \cite{ke2020smart} \cite{liu2018hybrid}& Individuals, Objects, Crowd, Automobiles \\
    & \cite{singh2018deep}  \cite{li2018deep}  \cite{ji20123d}  \cite{liu2016semi}   \cite{lai2019lstm} &  \\
    & \cite{li2018road}  \cite{liu2018accumulated}  \cite{zhao2020lightweight} \cite{deng2021air} \cite{zhai2020smart}\\
    & \cite{ma2019cost} \cite{xu2020trust} \\

    STONE & \cite{xu2019low} & Objects (Grassy Scene)\\
    AVIRIS & \cite{xu2019low} & Objects (Aerial View)\\
   SkyEye & \cite{roy2020detection} & Objects (Aerial View)\\
   UCF101 & \cite{roy2018unsupervised} & Individual (Action Videos) \\
   HMDB51 & \cite{roy2018unsupervised} & Individual (Action Videos)\\
   THUMOS14 & \cite{roy2018unsupervised} \cite{guo2018fully}  & Individual (Action Videos) \\
   BP4D & \cite{perveen2018spontaneous} & Individual (Facial Expressions)\\
   AFEW & \cite{perveen2018spontaneous} &  Individual (Facial Expressions)\\
   CAVIAR  & \cite{mo2013adaptive} & Individuals, Objects (Vehicles)\\
   PASCAL VOC& \cite{zhou2017fast} \cite{wang2019saliencygan} \cite{nikouei2018smart} \cite{nikouei2019toward}   & Objects (Vehicles)\\
   DS1, DS2 & \cite{muhammad2019efficient} & Events (Fire/Smoke/Fog)\\
   SFpark & \cite{bock2019smart} & Object (Taxis) \\
   LISA 2010 & \cite{zhou2017fast} & Objects (Vehicles)\\
   ImageNet & \cite{muhammad2019edge} & Events (Fire/Smoke/Fog)\\
   QMUL Junction & \cite{isupova2016anomaly} \cite{cheng2015gaussian} & Objects (Vehicles)\\
   HockeyFight & \cite{asad2021multi} & Individuals \\
   ViolentFlow & \cite{asad2021multi} & Individuals \\
   
   \hline

\end{tabular}
\end{table}

\subsection{Edge Computing in Anomaly Detection: Challenges and Opportunities}

\subsubsection{Challenge 1: Locating Edge Nodes}
Locating edge nodes in a distributed computing paradigm is well explored through a variety of techniques \cite{de2011toward} \cite{povedano2013dargos} \cite{montes2013gmone}. Bench-marking techniques are used for mapping tasks onto the most suitable resources and thereby discovering edge nodes. However, a proper mechanism is required to explore the edge of the network as several devices from different generations will be available at this layer. For example, machine learning tasks were rarely used previously but nowadays, they are used as a first-hand option for anomaly detection setups. Bench-marking methods should be efficient in finding the capability and availability of resources.
A diagram illustrating the challenges and opportunities is given in Figure 6.

\subsubsection{Challenge 2: Security and accessibility}
In edge computing, a significantly large number of nodes (edge devices) participate in processing tasks and each device requires a user access encryption \cite{kang2017privacy}. Also, the data that is processed needs to be secured as it is handled by many devices during the process of offloading \cite{guo2019detecting}. 

\subsubsection{Challenge 3: Quality of Service}
The quality of service delivered by the edge nodes is determined by the throughput where the aim is to ensure that the nodes achieve high throughput while delivering workloads. The overall framework should ensure that the nodes are not overloaded with work however if they are overloaded in the peak hours, the tasks should be partitioned and scheduled accordingly \cite{rodrigues2016hybrid} \cite{suto2015qoe}. 

\begin{figure}
    \centering
    \includegraphics[width=7cm]{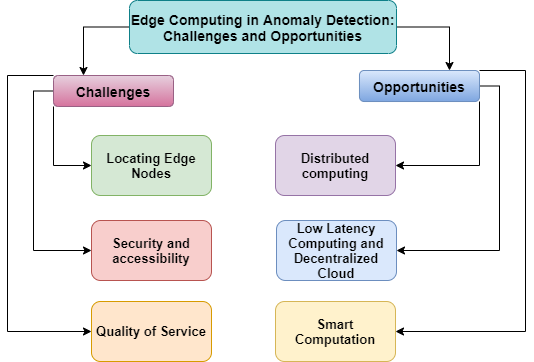}\\
    \caption{Edge Computing in Anomaly Detection: Challenges and Opportunities}
\end{figure}

\subsubsection{Opportunity 1: Distributed computing}
Edge computing uses the technique of dividing computationally expensive tasks to other nodes available in the network thereby reducing response time. The transfer of these intensive tasks to a separate processor such as a cluster, cloud-let, or grid is called computation offloading. It is used to accelerate applications by dividing the tasks between the nodes such as mobile devices. Mobile devices have physical limitations and are restricted in memory, battery, and processing. This is the reason that many computationally heavy applications do not run on such devices. To cope with this problem, the anomaly detection task is migrated to various edge devices according to the computing capabilities of respective devices. Xu et al. tried to optimize running performance, responsive time, and privacy by deploying task offloading for video surveillance in edge computing enabled Internet of Vehicles \cite{xu2020trust}. Similarly, a Fog-enabled real-time traffic management system uses a resource management offloading system to minimize the average response time of the traffic management server \cite{wang2018offloading} \cite{dinh2018learning} \cite{yang2015cost}. The resources are efficiently managed with the help of distributed computing or task offloading \cite{wang2018offloading}  \cite{zhao2020lightweight} \cite{wang2018traffic}.

\subsubsection{Opportunity 2: Low Latency Computing and Decentralized Cloud}
As far as anomaly detection using the cloud, the data is captured on the device and is processed away from the device leading to a delay. Moreover, if the cloud centers are geographically distant the time response is hampered further. Edge computing has the capability of processing the data where it is produced thereby reducing the latency \cite{nikouei2018smart} \cite{yang2019edge}.
Other conventional methods focused on improving either transmission delay or processing delay, but not both. Service delay puts forth a solution that reduces both \cite{rodrigues2016hybrid}.

\subsubsection{Opportunity 3: Smart Computation}
To perform meaningful analytics, the data generated at the user end needs to be transported to the cloud. There are inevitable delays and energy implications in this process. Computations can be performed smartly by hierarchically distributing the computations \cite{li2018deep}. Smart computation involves accuracy \cite{nikouei2019toward}, efficiency  \cite{xu2019data}, lower computation \cite{xu2018real}, latency \cite{wang2019saliencygan} which is essential for abnormal event detection. If there are limited resources on edge nodes, data centers offload the task to volunteer nodes to enhance computational capabilities of the front-end devices \cite{lai2019lstm}.

\section{Observations}
After studying different paradigms of anomaly detection in video surveillance systems, we observe that only benchmark data-set-based comparison may not be relevant for all real-life situations, as they are not enough to consider all real-life scenarios. Further, the performance depends on the density of the crowd, as the crowd increases the performance of the anomaly detection model decreases and it works best when the crowd is sparse. Some approaches intend to neglect background and focus only on foreground features for anomaly detection. We think that background information would be useful to model environmental conditions like rainy, sunny, or snowy weather that can cause anomalies. Further, for delay-sensitive applications like intelligence surveillance and anomaly detection, edge computing is a promising approach. It offers more privacy and security as the data is processed on the device itself. With continuous improvement in edge devices and task offloading the workload is divided thereby improving the overall efficiency.

\section{Conclusion}
In this paper, we survey anomaly detection in video surveillance. We explored various anomaly detection techniques applied for different surveillance scenarios including vehicular pedestrian, crowd, traffic, industries, and public places. We emphasized the learning techniques, models, approaches, and different scenarios for anomaly detection. The survey intended to provide detailed inside and diversities in anomaly detection techniques. In context to anomaly detection using edge computing, the area is less explored and needs attention.  A lot of work can be done in this field to achieve state-of-the-art anomaly detection and intelligence surveillance on edge devices.

\bibliographystyle{ieeetr}
\bibliography{main.bib}

\end{document}